\documentclass[conference]{IEEEtran}

\IEEEoverridecommandlockouts
\usepackage{cite}
\usepackage{amsmath,amssymb,amsfonts}
\usepackage{algorithmic}
\usepackage{graphicx}
\usepackage{textcomp}

\usepackage[inline]{enumitem}
\usepackage{xcolor}

\usepackage{pgfplots}
\pgfplotsset{compat=1.18}

\usepackage{witharrows}

\usepackage[flushleft]{threeparttable}
\usepackage{tablefootnote}
\usepackage{array}
\newcolumntype{P}[1]{>{\centering\arraybackslash}m{#1}}
\usepackage{multicol}
\usepackage{multirow}
\usepackage{tabulary}
\usepackage{colortbl}
\definecolor{mygray}{gray}{0.90}
\usepackage{makecell}
\usepackage{booktabs}
\usepackage{subcaption}
\usepackage{stmaryrd}
\usepackage{float}
\usepackage{pifont}

\usepackage{arydshln}
\colorlet{mygray}{gray!15!white}

\usepackage[switch,columnwise]{lineno}

\usepackage{url,hyperref,microtype}
\hypersetup{
    colorlinks=true,
    citecolor=blue,
    linkcolor=blue,
    filecolor=magenta,      
    urlcolor=black,
}

\def\BibTeX{{\rm B\kern-.05em{\sc i\kern-.025em b}\kern-.08em
    T\kern-.1667em\lower.7ex\hbox{E}\kern-.125emX}}

\begin{document}

\title{An Exploratory Analysis of Pain Localization via Explainable Computational Modeling \\
}

%%%%%%%%%%%%%%%%%%%%%%%%%%%%%%%%%%%%%%%%%%%%%%%

\author{

\IEEEauthorblockN{Ioannis Kyprakis}
\IEEEauthorblockA{\textit{Hellenic Mediterranean University} \\
Heraklion, Greece \\
ddk366@edu.hmu.gr}

\and

\IEEEauthorblockN{Stefanos Gkikas}
\IEEEauthorblockA{\textit{Honda Research Institute Japan} \\
Wako City, Japan \\
stefanos.gkikas@jp.honda-ri.com}

\and

\IEEEauthorblockN{Eric Nichols}
\IEEEauthorblockA{\textit{Honda Research Institute Japan} \\
Wako City, Japan \\
e.nichols@jp.honda-ri.com}

\and

\IEEEauthorblockN{Yu Fang}
\IEEEauthorblockA{\textit{Honda Research Institute Japan} \\
Wako City, Japan \\
yu.fang@jp.honda-ri.com}

\and

\IEEEauthorblockN{Manolis Tsiknakis}
\IEEEauthorblockA{\textit{Hellenic Mediterranean University} \\
Heraklion, Greece \\
tsiknaki@hmu.gr}

}

%%%%%%%%%%%%%%%%%%%%%%%%%%%%%%%%%%%%%%%%%%%%%%%%%%%%%%%%%%
\maketitle

\begin{abstract}
Automatic pain localization, which involves identifying the anatomical origin of pain from peripheral physiological signals without patient self-report, is a clinically critical but largely unaddressed problem, particularly for non-verbal patients. This paper presents a systematic comparison of classical feature engineering and deep sequence learning for subject-independent three-class pain localization using the AI4Pain 2026 Challenge dataset, which comprises four synchronously recorded wearable modalities: electrodermal activity, blood volume pulse, respiration, and peripheral oxygen saturation recorded from 65 participants under controlled TENS-induced pain. A 115-dimensional hand-crafted feature set spanning time-domain, frequency-domain, modality-specific, and cross-modal descriptors is benchmarked against end-to-end deep architectures. Extremely Randomized Trees achieves the highest macro-F1 of 0.539, outperforming the best deep model by 7.4 percentage points, with EDA spectral features emerging as the dominant discriminators. A consistent 26-point gap between pain detection (F1\,=\,0.815) and localization (F1\,=\,0.552) across all models points to a fundamental ceiling imposed by the anatomical diffuseness of peripheral autonomic pathways at 10-second resolution.
\end{abstract}


\section{Introduction}

Pain plays a fundamental evolutionary role, serving as a warning mechanism that alerts organisms to potential bodily harm or disease and helps preserve physical integrity \cite{santiago_2022}. It is inherently subjective and multidimensional, encompassing nociceptive, sensory, affective, and cognitive dimensions \cite{marchand_2024}. Given its widespread societal burden, pain has been characterized as a \textit{\textquotedblleft silent public health epidemic\textquotedblright} \cite{katzman_gallagher_2024}. Its considerable global prevalence, affecting more than 30\% of the world's population, places an enormous personal and economic burden on individuals and healthcare systems alike \cite{cohen_vase_2021}. Opioids have historically represented one of the most widely prescribed analgesic interventions \cite{kaye_jones_2017}; however, their use carries considerable risks, including dependency, abuse, and overdose \cite{stampas_pedroza_2020}, increased stress levels \cite{kassiotis_stressgat_acii_2026}, as well as side effects that degrade daily functioning and reduce overall quality of life \cite{benyamin_trescot_2008}. The evaluation of pain remains a complex task. Current clinical practice relies primarily on patient self-reports, which offer limited insight into treatment efficacy and also have been linked to the overprescription and misuse of analgesics \cite{kong_chon_2024}. Patients who are acutely or chronically ill, or who lack the capacity for clear verbal expression, face a heightened risk of inadequately treated pain \cite{herr_coyne_2019}.

In adult critical care settings, pain management remains a constant clinical challenge, with established guidelines emphasizing the need for standardized assessment protocols to support appropriate intervention decisions. This challenge is further intensified in individuals with cancer-related pain, particularly those approaching end-of-life \cite{snijders_brom_2022}. Beyond its physical dimensions, chronic pain imposes a substantial cognitive burden, impairing attentional capacity, reducing functional performance, and prolonging reaction times \cite{khera_rangasamy_2021}. Chronic pain has further been shown to impact cognitive control mechanisms, including expectancy processing and decision-making \cite{li_lyu_2025}. In addition, both chronic and acute pain influence attentional performance \cite{moore_meints_2019}.

Beyond detecting the presence or intensity of pain, identifying its anatomical origin is a clinically critical yet underexplored challenge. Knowing where pain originates enables targeted intervention and informs diagnostic decisions in ways that intensity scores alone cannot provide \cite{jaatun_hjermstad_2014}. This is particularly significant for non-verbal patients, who are unable not only to quantify their pain but also to indicate its location, leaving clinicians without the spatial information necessary for appropriate care. For example, it has been shown that sedation may conceal uncontrolled pain in intubated patients while simultaneously preventing them from communicating their discomfort to nursing staff \cite{clukey_weyant_2014}. Among mechanically ventilated patients, over half are estimated to experience moderate to severe unrelieved pain \cite{ayasrah_2019}, often with no reliable means of communicating its source. Despite growing recognition of pain localization as a key objective for automated pain recognition systems \cite{walter_gruss_frisch_2020}, no clinically validated solution currently exists for non-verbal patients. Consequently, a system capable of both objectively detecting pain and localizing its source would offer substantial clinical value, whether in hospital settings or in home-based monitoring.

\section{Related Work}
\label{related_work}
Over the last decade, research on computational pain assessment has expanded considerably. While deep learning-based models have become the dominant paradigm, traditional methods continue to deliver strong, competitive results, particularly when labeled data are limited \cite{gkikas_phd_thesis_2025}. The majority of existing approaches rely on video data, capturing behavioral manifestations of pain through facial expressions, body motion, and other observable cues across a range of learning strategies \cite{gkikas_reface_2026,gkikas_tsiknakis_embc}. Although video-based methods dominate the field, a growing body of work has explored biosignal-driven approaches to pain recognition. Prior studies have drawn on physiological markers from electrocardiography \cite{gkikas_chatzaki_2023}, electromyography \cite{thiam_bellmann_kestler_2019,werner_hamadi_niese_2014,patil_patil_2024,pavlidou_tsiknakis_2025}, electrodermal activity (EDA) \cite{gkikas_kyprakis_eda_2025,phan_iyortsuun_2023,lu_ozek_kamarthi_2023,li_luo_2024,aziz_joseph_2025}, respiration rate\cite{gkikas_kyprakis_resp_2025}, and brain activity recorded via functional near-infrared spectroscopy \cite{rojas_huang_2016,rojas_liao_2019,rojas_romero_2021,khan_sousani_2024,rojas_joseph_bargshady_2024,gkikas_arzate_eeite_pain_2026, bargshady_aziz_2025}. For broader overviews of pain recognition frameworks and evidence across modalities, we refer to \cite{gkikas_tsiknakis_slr_2023,khan_umar_2025}. Multimodal approaches have also attracted considerable interest, with studies showing that combining behavioral and physiological information can improve robustness and recognition performance \cite{farmani_bargshady_2025, gkikas_arzate_pain_icmi_2026, gkikas_tiny_2025}. In \cite{shi_chikhaoui_2022}, the authors fused ECG and EDA features using tree-based classifiers, while in \cite{jerritta_murugappan_2022}, wavelet-based ECG and EMG features were combined with a k-nearest-neighbor classifier. Multimodal fusion has also been applied in pediatric postoperative settings, where \cite{liang_luo_2025} demonstrated gains from integrating facial and voice features. Beyond the laboratory, the authors in \cite{badura_2021} examined pain during physiotherapy sessions using a setup that included EDA, EMG, respiration, blood volume pulse (BVP), and hand force. Other work has focused on jointly modeling physiological and behavioral signals: the authors in \cite{zhi_yu_2019} combined facial videos with ECG, EMG, and EDA by extracting and fusing handcrafted descriptors, while \cite{gkikas_tachos_2024} integrated facial videos with heart-rate information within a transformer-based framework for pain intensity estimation. A separate line of work extracts pseudo-physiological signals directly from facial video and fuses them with the original visual features; in \cite{huang_dong_2022}, 3D CNNs were used to estimate pseudo-heart-rate signals from facial video.

Despite the growing body of work on automatic pain assessment, the problem of identifying the anatomical source of pain has received little attention \cite{walter_gruss_frisch_2020}. In \cite{aziz_khan_2023}, the authors used EDA signals to simultaneously detect and localize acute pain, framing localization as a classification task over distinct body regions. Building on this, \cite{aziz_joseph_2025} proposed a two-stage architecture that introduced multi-domain binary pattern features extracted from EDA. A complementary line of work has approached pain localization through computational analysis of patient-reported body maps. In \cite{correa_bittencourt_2020}, the authors developed the PainMAP software for automated quantification of pain drawings on body charts, while \cite{abudawood_yoon_2023} proposed an algorithmic method for extracting spatial pain coordinates from digital manikins and applied it to patients with sickle cell disease.
This work explores multiple physiological modalities for automatic pain localization, extracts and evaluates modality-specific features, and provides an interpretable analysis of their relative contributions to the localization task.

\section{Methodology}
\label{sec:methodology}

% ------------------------------------------------------------------
\subsection{Dataset}
\label{ssec:data}

\subsubsection{Participants}
The AI4Pain dataset was collected at the Human-Machine Interface Laboratory at the University of Canberra, Australia~\cite{ai4pain2026, ai4pain2025, ai4pain2024, fernandez_2023_multimodal}. It comprises recordings from 65 participants (23 female, mean age $29.06 \pm 8.28$ years). Participants were seated comfortably with both arms resting on a table throughout the session.

Four modalities were recorded synchronously at 100\, Hz using Biosignal Plux wearable sensors:
EDA, reflecting sympathetic nervous system arousal via galvanic skin response. BVP, capturing cardiac pulse morphology and heart rate via photoplethysmography. Resp, encoding breathing rate and depth via chest expansion. Peripheral Oxygen Saturation (SpO\textbf{\textsubscript{2}}), measured via pulse oximetry.

A balanced dataset was constructed with three classes: No Pain, Pain Arm, and Pain Hand. Each subject contributes exactly 12 segments per class ($\approx$10\,s each), for a total of 36 segments per subject.

The dataset is divided into three subject-disjoint splits: training (41 subjects, 1{,}476 segments), validation (12 subjects, 432 segments), and test (12 subjects, 432 segments). All three splits are perfectly class-balanced at 33.3\% per class. The subject-independent partitioning ensures no participant's physiological baseline appears across splits, enforcing generalization across individuals.

% ------------------------------------------------------------------
\subsection{Preprocessing and Feature Extraction}
\label{ssec:preproc}

\subsubsection{Signal Preprocessing}
Raw signals are processed through a modality-specific filtering pipeline using zero-phase Butterworth filters to suppress motion artifacts, baseline drift, and high-frequency noise. BVP is bandpass filtered at 0.5--5.0\, Hz (4th-order Butterworth). EDA and Respiration are low-pass filtered at 1.0\, Hz (4th-order  Butterworth). SpO\textsubscript{2} is first clipped to the physiological range $[80, 100]$\% to remove sensor artifacts and then smoothed with a 0.5\, Hz low-pass filter (2nd-order Butterworth). Each channel is then independently z-score normalized using statistics computed exclusively from the training split, ensuring no information leakage to the validation or test sets.
% All segments are zero-padded or truncated to a fixed length of 1{,}000 samples (10\,s at 100\, Hz).

\subsubsection{Feature Extraction}
We extract 115 hand-crafted features per segment, organized
into four groups:

\paragraph{Time-domain features (64 features, 16 per channel)}
For each modality: mean, median, standard deviation, variance, minimum, maximum, range, interquartile range, skewness, kurtosis, root mean square, linear slope, area under the curve (trapezoidal rule), zero-crossing rate, signal energy, and sample entropy.

\paragraph{Frequency-domain features (24 features, 6 per channel)}
Using the Welch periodogram: dominant frequency, spectral entropy, total power, and power in three physiologically motivated frequency bands (low/mid/high) with modality-specific band boundaries.

\paragraph{Modality-specific features (21 features)}

\textit{BVP (9)}: peak count, estimated heart rate (bpm), pulse amplitude mean and standard deviation, inter-beat interval (IBI) mean, standard deviation, and RMSSD (an HRV index), and spectral power in 0.5–2\, Hz and 2–5\, Hz bands.

\textit{EDA (6)}: tonic component mean (skin conductance level, SCL); phasic component standard deviation and maximum (skin conductance response, SCR amplitude); number of SCR peaks; mean and maximum SCR peak amplitudes.

\textit{Resp (4)}: estimated respiration rate (breaths/min), mean and standard deviation of cycle amplitude, and cycle-to-cycle variability.

\textit{SpO\textsubscript{2} (2)}: number of desaturation events and percentage of samples below 95\%.

\paragraph{Cross-modal correlation features (6 features)}
Pearson correlation coefficients between all six modality pairs: BVP$\times$EDA, BVP$\times$Resp, BVP$\times$SpO\textsubscript{2}, EDA$\times$Resp, EDA$\times$SpO\textsubscript{2}, and Resp$\times$SpO\textsubscript{2}.

% ------------------------------------------------------------------
% ------------------------------------------------------------------
\subsection{Models}
\label{ssec:models}

\subsubsection{Classical Feature-Based Methods}
\label{sssec:classical}

All classical models operate on the 115-dimensional feature vector described in Section~\ref{ssec:preproc} and are preceded by z-score standardization. Hyperparameter optimization is performed via randomized search with 200 iterations and 5-fold cross-validation stratified by subject identity, ensuring that no participant appears simultaneously in the training folds and the held-out fold.

\paragraph{Extra Trees}
Our primary classifier is the Extremely Randomized Trees algorithm~\cite{geurts2006}, which extends the Random Forest by additionally randomizing split thresholds, thereby reducing variance at the cost of a small increase in bias. The final model comprises 100 trees with a maximum depth of 10, a minimum of one sample per leaf, square-root feature subsampling, and balanced class weights to account for between-subject variability in segment counts. A key motivation for selecting this architecture is its ability to compute Gini impurity-based feature importances at no additional inference cost, thereby enabling direct post hoc explainability of the learned pain representation.

\paragraph{Ensemble}
Soft-voting ensembles were constructed by combining the probabilistic outputs of six classical classifiers: Logistic Regression, Support Vector Machine with radial basis function kernel (RBF-SVM), Random Forest, Extra Trees, XGBoost, and LightGBM. Each classifier was trained on the same 115-dimensional handcrafted feature representation. Class posterior probabilities were averaged to obtain the final prediction. Two ensemble variants were evaluated: an equal-weight ensemble, in which all classifiers contributed equally, and an optimized-weight ensemble, in which the convex combination coefficients were selected through grid search to maximize macro-averaged F1 on the validation set.

\subsubsection{Deep Sequence Models}
\label{sssec:deep}

All deep architectures receive the preprocessed four-channel signal of shape $4 \times 1{,}000$ as direct input, without any hand-crafted feature extraction. A unified training protocol is applied across all architectures: the AdamW optimizer with a warmup--cosine--cooldown learning-rate schedule (10\% linear warm-up, 80\% cosine annealing, 10\% cosine cool-down to 5\% of the peak rate), cross-entropy loss with label smoothing ($\varepsilon = 0.05$), and patience-based early stopping after 30 non-improving epochs. To mitigate class imbalance at the mini-batch level, training samples are drawn with probabilities proportional to their class frequencies. Light stochastic augmentation is applied at each iteration: additive Gaussian noise with standard deviation $\sigma = 0.02$, independent per-channel amplitude scaling uniformly sampled from $[0.85, 1.15]$.

\paragraph{CNN-Transformer}
The signal is first passed through a 1-D convolutional stem comprising three successive convolution--batch normalization--ReLU blocks with progressively increasing dilation, which captures local temporal patterns while downsampling the sequence to 125 time steps. A single Transformer encoder layer with pre-layer normalization and multi-head self-attention then aggregates global dependencies across the compressed sequence. A classification token prepended to the sequence accumulates the global representation, which is subsequently projected to class logits by a two-layer feed-forward head.

\paragraph{Fusion Network}
To explicitly account for the heterogeneous nature of the four physiological modalities, the Fusion Network employs dedicated convolutional encoders for each signal channel. The per-modality embeddings are aggregated via a learned attention gate that produces a scalar importance weight for each channel, enabling the network to dynamically up-weight informative modalities on a per-sample basis. The weighted concatenation is passed through a shared multi-layer perceptron to produce the final class prediction.

Table~\ref{tab:models_summary} summarizes the key properties of all evaluated architectures.

\begin{table}[t]
\centering
\caption{Summary of evaluated model architectures.
$^\dagger$Parameter counts are approximate for models without
hyperparameter search.}
\label{tab:models_summary}
\resizebox{\columnwidth}{!}{%
\begin{tabular}{llccc}
\toprule
\textbf{Type} & \textbf{Model} & \textbf{Input} & \textbf{Parameters} & \textbf{Optimisation} \\
\midrule
\multirow{2}{*}{Classical}
  & Extra Trees  & 115 features & ---  & 200-iter randomised search \\
  & Ensemble     & 115 features & ---  & Grid search (validation set) \\
\midrule
\multirow{2}{*}{Deep}
  & CNN-Transformer & $4\times1{,}000$ & ${\sim}200$k$^\dagger$ & Fixed architecture \\
  & Fusion Network  & $4\times1{,}000$ & ${\sim}120$k$^\dagger$ & Fixed architecture \\
\bottomrule
\end{tabular}}
\end{table}

\section{Results}
\label{sec:results}

\subsection{Overall Classification Performance}
\label{ssec:overall}

Table~\ref{tab:main} reports macro-averaged F1, balanced accuracy, and one-vs-rest ROC-AUC on the held-out validation set for all evaluated models. The primary evaluation metric is macro-averaged F1, which weights each of the three classes equally.

Classical feature-based methods consistently outperform deep sequence models across all reported metrics. Extra Trees achieves the highest macro-F1 of 0.539 and a balanced accuracy of 0.549, representing a 7.4 percentage point improvement over the best deep model (CNN-Transformer, F1\,=\,0.465). Among classical methods, the optimized soft-voting ensemble (F1\,=\,0.535) ranks second, offering only marginal improvement over the single Extra Trees model, suggesting that constituent classifiers share correlated error patterns.

Among deep architectures, the CNN-Transformer achieves the highest macro-F1 (0.465), substantially outperforming the Fusion Network (0.413) despite the latter's modality-specific encoder design. The ROC-AUC gap between paradigms is pronounced: Extra Trees achieves 0.714 compared to 0.644 for CNN-Transformer.

\begin{table}[!h]
\centering
\caption{Validation-set performance of all evaluated models,
         ranked by macro-averaged F1.
         }
\label{tab:main}
\resizebox{\columnwidth}{!}{%
\begin{tabular}{llccc}
\toprule
\textbf{Type} & \textbf{Model}
  & \textbf{Macro-F1} & \textbf{Bal-Acc} & \textbf{ROC-AUC} \\
\midrule
\multirow{3}{*}{\rotatebox[origin=c]{90}{\hspace{5pt}Classical}}
  & Extra Trees              & \textbf{0.539} & \textbf{0.549} & 0.714 \\
  & Ensemble (optimised)     & 0.535          & 0.539          & \textbf{0.722} \\
  & Ensemble (equal weight)  & 0.532          & 0.537          & 0.729 \\
\midrule
\multirow{2}{*}{\rotatebox[origin=c]{90}{Deep}}
  & CNN-Transformer          & 0.465          & 0.465          & 0.644 \\
  & Fusion Network           & 0.413          & 0.414          & 0.597 \\
\bottomrule
\end{tabular}}
\end{table}

\subsection{Per-Class Analysis}
\label{ssec:perclass}

Table~\ref{tab:perclass} presents precision, recall, and F1 for each pain class for the three highest-performing models. All models identify the No Pain class most reliably, with recall values between 0.569 and 0.778, while the two pain localization classes (PainArm and PainHand) are substantially harder, with recall consistently below 0.47.

Extra Trees achieves the highest No Pain recall (0.778), but at the cost of lower recall on the two pain classes (0.438 and 0.431 for PainArm and PainHand). Arm and Hand pain classes show near-identical precision and recall across all models, confirming that the two localization classes pose a comparable, symmetrical challenge.

To quantify this asymmetry, we derive two secondary metrics directly from the 3-class model's predictions: a pain-detection F1, obtained by collapsing PainArm and PainHand into a single Pain class, and a localization F1, evaluated only on pain-positive segments (PainArm vs.\ PainHand). Extra Trees achieves a pain-detection F1 of 0.815 but a localization F1 of only 0.552, a 26-percentage-point disparity, as discussed in Section~\ref{sec:discussion}.

\begin{table}[t]
\centering
\caption{Per-class precision (P), recall (R), and F1 on the
         validation set for the three highest-performing models.}
\label{tab:perclass}
\resizebox{\columnwidth}{!}{%
\begin{tabular}{lccccccccc}
\toprule
& \multicolumn{3}{c}{\textbf{No Pain}}
& \multicolumn{3}{c}{\textbf{Pain Arm}}
& \multicolumn{3}{c}{\textbf{Pain Hand}} \\
\cmidrule(lr){2-4}\cmidrule(lr){5-7}\cmidrule(lr){8-10}
\textbf{Model}
  & P & R & F1 & P & R & F1 & P & R & F1 \\
\midrule
Extra Trees
  & .622 & .778 & .691
  & .496 & .438 & .465
  & .496 & .431 & .461 \\
Ensemble (optimized)
  & .658 & .708 & .682
  & .465 & .514 & .488
  & .483 & .396 & .388 \\
CNN-Transformer
  & .582 & .569 & .575
  & .407 & .458 & .431
  & .411 & .368 & .388 \\
\bottomrule
\end{tabular}}
\end{table}

\subsection{Modality Ablation}
\label{ssec:ablation}

To quantify the individual and joint contribution of each physiological signal source, we evaluate the primary Extra Trees classifier and CNN-Transformer across all single-modality subsets, pairwise combinations, and three-modality subsets. Table~\ref{tab:ablation} reports the single-modality results, the best-performing pair (BVP$+$EDA), and the best-performing triple (BVP$+$EDA$+$Resp). Cross-modal correlation features are included only in the all-four-modality condition, as partial subsets do not admit a complete pairwise correlation set.

EDA is the most informative single modality for both classical (F1\,=\,0.472) and deep (F1\,=\,0.430) models, substantially exceeding BVP (0.437 and 0.413) and rendering Resp and SpO\textsubscript{2} near chance level when used alone. Adding BVP to EDA yields the largest marginal gain for Extra Trees ($+$0.040 F1), while the deep model shows a negligible benefit ($+$0.014), suggesting the CNN-Transformer does not effectively exploit cardiac features encoded in the hand-crafted BVP representation. The BVP$+$EDA combination (F1\,=\,0.512) recovers 95\% of full four-modality Extra Trees performance (0.539), confirming these two channels as the core sensor set at 10-second resolution. Notably, adding Resp to BVP$+$EDA slightly reduces Extra Trees performance ($-$0.015), while the full four-modality set recovers an additional $+$0.042 F1 driven primarily by the six cross-modal correlation features and SpO\textsubscript{2}, which together encode co-modulation patterns unavailable in any individual channel subset.

\begin{table}[!h]
\centering
\caption{Modality ablation: macro-F1 for Extra Trees and
         CNN-Transformer on restricted feature subsets.}
\label{tab:ablation}
\begin{tabular}{lccc}
\toprule
\textbf{Modality subset}
  & \textbf{\# features}
  & \textbf{Extra Trees}
  & \textbf{CNN-Transformer} \\
\midrule
EDA only                  & 28  & 0.472 & 0.430 \\
BVP only                  & 31  & 0.437 & 0.413 \\
Resp only                 & 26  & 0.376 & 0.382 \\
SpO\textsubscript{2} only & 24  & 0.370 & 0.304 \\
BVP $+$ EDA               & 59  & 0.512 & 0.416 \\
BVP $+$ EDA $+$ Resp      & 85  & 0.497 & 0.412 \\
All four modalities        & 115 & \textbf{0.539} & \textbf{0.465} \\
\bottomrule
\end{tabular}
\end{table}

\subsection{Feature Importance and Explainability}
\label{ssec:explainability}

Gini impurity-based importance scores are extracted from the trained Extra Trees model at no additional computational cost. At the modality level, EDA accounts for 37.9\% of aggregate importance, followed by BVP (26.7\%), Respiration (16.7\%), and SpO\textsubscript{2} (14.5\%). The remaining 4.2\% is distributed across the six cross-modal correlation features. This ranking is fully consistent with the ablation results.

\begin{figure}[!h]
\centering
\begin{tikzpicture}
\begin{axis}[
    xbar,
    width=0.95\columnwidth,
    height=4.8cm,
    xmin=0,
    xmax=3.6,
    xlabel={Importance (\%)},
    xlabel style={font=\scriptsize},
    xticklabel style={font=\scriptsize},
    ytick={1,2,3,4,5,6,7,8,9,10},
    yticklabels={
        EDA dom. freq.,
        EDA spec. ent.,
        EDA kurtosis,
        EDA skewness,
        BVP samp. ent.,
        BVP spec. ent.,
        EDA mean,
        Resp spec. ent.,
        BVP RMSSD,
        EDA low-band
    },
    y dir=reverse,
    yticklabel style={font=\tiny},
    tick label style={font=\tiny},
    bar width=4pt,
    enlarge y limits=0.08,
    nodes near coords,
    point meta=x,
    every node near coord/.append style={
        font=\tiny,
        anchor=west,
        xshift=1pt
    },
    axis x line*=bottom,
    axis y line*=left,
    grid=major,
    major grid style={draw=gray!20},
    clip=false
]
\addplot coordinates {
    (3.21,1)
    (2.98,2)
    (2.74,3)
    (2.61,4)
    (2.43,5)
    (2.31,6)
    (2.18,7)
    (2.05,8)
    (1.97,9)
    (1.89,10)
};
\end{axis}
\end{tikzpicture}
\caption{Top-10 features ranked by Gini importance.}
\label{fig:importance}
\end{figure}
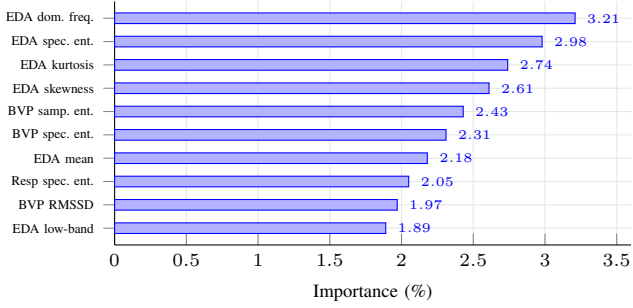

Figure~\ref{fig:importance} lists the ten highest-ranked individual features. EDA spectral descriptors occupy the top four positions: dominant frequency (3.21\%), spectral entropy (2.98\%), kurtosis (2.74\%), and skewness (2.61\%). BVP features appear at ranks 5 and 6 via sample entropy (2.43\%) and spectral entropy (2.31\%), while BVP RMSSD ranks ninth (1.97\%). No SpO\textsubscript{2} feature appears among the top-ranked individual features.

A secondary analysis restricted to the two pain classes alone reveals a shift in the discriminative feature set: BVP inter-beat interval variability (RMSSD and IBI standard deviation) and EDA phasic characteristics: maximum SCR amplitude and SCR peak count, rise in relative importance, indicating that pain localization relies more on the magnitude and variability of the autonomic response than on its spectral structure.

\subsection{Test Set Performance}
\label{ssec:test}

Following the release of the held-out test set by the 
challenge organizers, the Extra Trees model selected on 
the basis of validation macro-F1 was evaluated under the 
official challenge protocol, which reports plain accuracy. 
The model achieves a test set accuracy of 49.1\%, compared 
to a validation balanced accuracy of 54.9\%. Given that 
all three splits are perfectly class-balanced, plain 
accuracy and balanced accuracy are numerically equivalent, 
making the two metrics directly comparable. The 5.8 
percentage point reduction from validation to test is 
consistent with the inter-subject variability expected 
in a subject-independent evaluation with only 12 test 
subjects per split.

%OR USE TABLE BELOW
% \begin{table}[!h] \centering \caption{Top-10 individual features ranked by Gini importance from the Extra Trees classifier.} \label{tab:importance} \begin{tabular}{clcc} \toprule \textbf{Rank} & \textbf{Feature} & \textbf{Modality} & \textbf{Importance (\%)} \\ \midrule 1 & EDA dominant frequency & EDA & 3.21 \\ 2 & EDA spectral entropy & EDA & 2.98 \\ 3 & EDA kurtosis & EDA & 2.74 \\ 4 & EDA skewness & EDA & 2.61 \\ 5 & BVP sample entropy & BVP & 2.43 \\ 6 & BVP spectral entropy & BVP & 2.31 \\ 7 & EDA mean & EDA & 2.18 \\ 8 & Resp spectral entropy & Resp & 2.05 \\ 9 & BVP RMSSD & BVP & 1.97 \\ 10 & EDA low-band power & EDA & 1.89 \\ \bottomrule \end{tabular} \end{table}

% ─────────────────────────────────────────────────────────────
% DISCUSSION
% ─────────────────────────────────────────────────────────────
\section{Discussion}
\label{sec:discussion}

The central finding of this study is that an interpretable classical machine learning pipeline substantially outperforms all evaluated deep sequence architectures on subject-independent pain localization. This result reflects the limitations of end-to-end temporal feature learning under the small-sample, short-segment conditions of wearable autonomic recordings, where domain-informed statistical representations provide a stronger inductive bias than patterns discovered directly from raw signals.

With $1{,}476$ training segments from $41$ subjects, deep models operating directly on raw $4 \times 1{,}000$ signals face a high-dimensional, small-sample setting in which the risk of learning subject-specific rather than pain-specific patterns is high. The $115$-dimensional, manually designed feature representation avoids this issue by encoding each segment as a set of physiologically inspired summary statistics, which are inherently insensitive to subject-specific variations in signal amplitude and baseline level. Critically, this advantage is not merely quantitative: the proposed approach produces a ranked list of discriminative features that directly reflect the underlying physiology, a property essential for clinical decision support.

The feature importance analysis provides a physiologically grounded explanation of how the model distinguishes pain states. EDA spectral features (dominant frequency and spectral entropy) emerge as the strongest individual discriminators, consistent with the well-established role of the sympathetic nervous system in mediating the skin conductance response to nociceptive stimuli \cite{papanikolaou2026}. Pain activates sympathetic preganglionic neurons that drive eccrine sweat gland secretion, producing characteristic shifts in EDA spectral content that reflect the timing and rhythm of sympathetic bursts rather than their gross amplitude. The prominence of spectral over amplitude features suggests that the temporal organization of the sympathetic response most reliably distinguishes pain conditions across subjects with heterogeneous baseline EDA levels. The secondary contribution of BVP entropy and IBI variability reflects the concurrent suppression of cardiac parasympathetic tone during acute nociception \cite{forte2022}. That entropy-based cardiac descriptors outrank conventional heart rate statistics further suggests that pain disrupts the regularity of cardiac autonomic modulation more consistently than its mean level.

Both the modality ablation and the importance analysis support the value of combining multiple physiological modalities. Although EDA and BVP together recover $95\%$ of full-system performance, the cross-modal correlation between these two signals contributes additional discriminative information that neither channel provides independently.

During sympathetic activation, BVP amplitude and EDA amplitude are co-modulated by the same autonomic drive, producing a characteristic covariation pattern that is reliably disrupted by pain. Respiration and SpO\textsubscript{2} contribute only marginally within $10$-second windows, most likely because their pain-relevant dynamics evolve over timescales that a single short segment cannot capture. This does not suggest these channels should be excluded; their limited contribution is specific to the current window length, and the discriminative value of each modality is inseparable from the temporal resolution at which it is observed. For respiration, pain-relevant changes, including alterations in respiratory rate, depth, and minute ventilation, unfold over multiple respiratory cycles and may require windows substantially longer than 10 seconds to characterize reliably \cite{jafari2017}. For SpO\textsubscript{2}, pain-relevant dynamics reflect cumulative cardiovascular and ventilatory changes that evolve too slowly to manifest within a single 10-second segment. Future work should therefore explore adaptive or multi-scale temporal windows that match the characteristic response time of each modality, rather than applying a single fixed segment length across all channels.

A further important finding concerns the asymmetry between pain detection and pain localization. All models find it substantially easier to identify the presence of pain than to distinguish its anatomical origin, with the best classifier achieving a pain detection F1 of $0.815$ alongside a localization F1 of $0.552$, a gap of $26$ percentage points. This disparity is physiologically expected: the sympathetic efferent pathways mediating EDA and cardiac responses to nociception are anatomically diffuse and are not organized by body region at the level of the peripheral sensors accessible via wearables. Arm and hand pain activate qualitatively similar autonomic responses that differ primarily in magnitude and latency rather than in spectral character or morphological pattern. The $26$-point detection--localization gap may therefore represent a performance ceiling for autonomic-only wearable classification at $10$-second resolution. This finding has direct clinical implications for the non-verbal patient scenario that motivates this work: while wearable autonomic signals can reliably detect the presence of pain, localizing its anatomical source from peripheral physiology alone remains an open problem and may require integrating complementary modalities, such as facial video or somatosensory mapping.

Several limitations should be noted. The validation set comprises only 12 subjects, making performance estimates subject to non-trivial sampling variability. The observed drop from validation balanced accuracy (54.9\%) to test set accuracy (49.1\%) further reflects this sensitivity to the specific physiological profiles of the subjects sampled in each split. Pain was induced using a TENS device under controlled laboratory conditions, and generalization to chronic, post-operative, or neuropathic pain remains an open question. Future work should explore hybrid architectures that preserve the interpretability demonstrated here while incorporating learned temporal representations, as well as longitudinal, multi-site datasets that cover the full range of clinical pain presentations.

% ─────────────────────────────────────────────────────────────
% CONCLUSION
% ─────────────────────────────────────────────────────────────
\section{Conclusion}
\label{sec:conclusion}

We presented a systematic comparison of classical and deep learning approaches for subject-independent pain localization in the AI4Pain 2026 Challenge. Extra Trees with $115$ hand-crafted physiological features achieves a validation macro-F1 of $0.539$, outperforming the best deep model by $7.4$ percentage points, confirming that domain-informed representations generalize more effectively than end-to-end learning in this small-sample regime. On the held-out test set, the model achieves an accuracy of 49.1\% under the official challenge evaluation protocol, confirming above-chance generalisation to fully unseen subjects. EDA dominates feature importance ($37.9\%$ Gini importance), with its spectral structure rather than gross amplitude serving as the most reliable discriminator across subjects. The consistent $26$-point gap between pain detection (F1\,=\,$0.815$) and localization (F1\,=\,$0.552$) reflects the anatomical diffuseness of peripheral sympathetic pathways and likely represents a performance ceiling for wearable autonomic classification at $10$-second resolution, pointing to the need for complementary modalities to fully address the pain localization problem in clinical practice.

\section*{Ethical Impact Statement}
All experiments are based on the \textit{AI4Pain} dataset, provided by the challenge organizers. Prior to collecting data, participants had no known histories of neurological or psychiatric disorders, unstable medical conditions, chronic pain, or regular medication use. On entry to the experiment, participants were given full details of the experimental procedure. Written informed consent was required as a condition for participation in this study. Approval for this study's protocol was received by the Human Ethics Committee of the University of Canberra (\textit{approval number: 11837}). This study and the proposed framework aim to assist in the continuous objective monitoring of pain and to reduce reliance on subjective clinical judgment. However, for practical application in clinical settings, further validation through dedicated clinical trials will be required before inclusion in standard care pathways. 

\section*{Acknowledgments}
The authors used large language model (LLM)-based tools for language editing and improvement. All scientific content, results, and conclusions are solely the work of the authors.

\bibliographystyle{IEEEtran}
\bibliography{library}

%\clearpage

\end{document}